\documentclass[letterpaper, 10 pt, conference]{ieeeconf}  

\IEEEoverridecommandlockouts                              
\usepackage{graphicx}
\usepackage{hyperref}
\usepackage[utf8]{inputenc}
\usepackage{url}
\usepackage{algorithm}
\usepackage{multirow}
\usepackage{algorithmic}
\usepackage{amssymb,amsmath}
\usepackage{graphicx}
\usepackage{subfigure}
\usepackage{multirow}
\usepackage{stfloats}
\usepackage{listings}

\usepackage{url}
\usepackage{algorithm}
\usepackage{multirow}
\usepackage{algorithmic}
\usepackage{amssymb,amsmath}
\usepackage{graphicx}
\usepackage{subfigure}
\usepackage{multirow}
\usepackage{stfloats}
\usepackage{listings}

\usepackage{tikz}
\usetikzlibrary{shapes,snakes}
\usepackage{amsmath,amssymb}
\title{\LARGE \bf
Towards a Robot Perception Specification Language
}
\author{Nico Hochgeschwender, Sven Schneider, Holger Voos, and Gerhard K. Kraetzschmar
\thanks{Nico Hochgeschwender, Sven Schneider, and Gerhard Kraetzschmar are with the Department of
        Computer Science, Bonn-Rhein-Sieg University of Applied Sciences,
        Germany. Email:
        {\tt\small firstname.lastname@h-brs.de}
        Nico Hochgeschwender and Holger Voos are with the Research Unit in Engineering Sciences,  University of Luxembourg, Luxembourg. Email:
        {\tt\small firstname.lastname@uni.lu}   
        }%
 }

\begin{document}

\maketitle
\thispagestyle{empty}
\pagestyle{empty}

\begin{abstract}
In this paper we present our work in progress towards a domain-specific language called \emph{Robot Perception Specification Language (RPSL)}. 
RSPL provide means to specify the expected result (task knowledge) of a Robot Perception Architecture in a declarative and framework-independent manner.

\end{abstract}
\section{Introduction}
\label{sec:intro}
\begin{figure}[ht]
  \centering
  \includegraphics[height=0.18\textwidth]{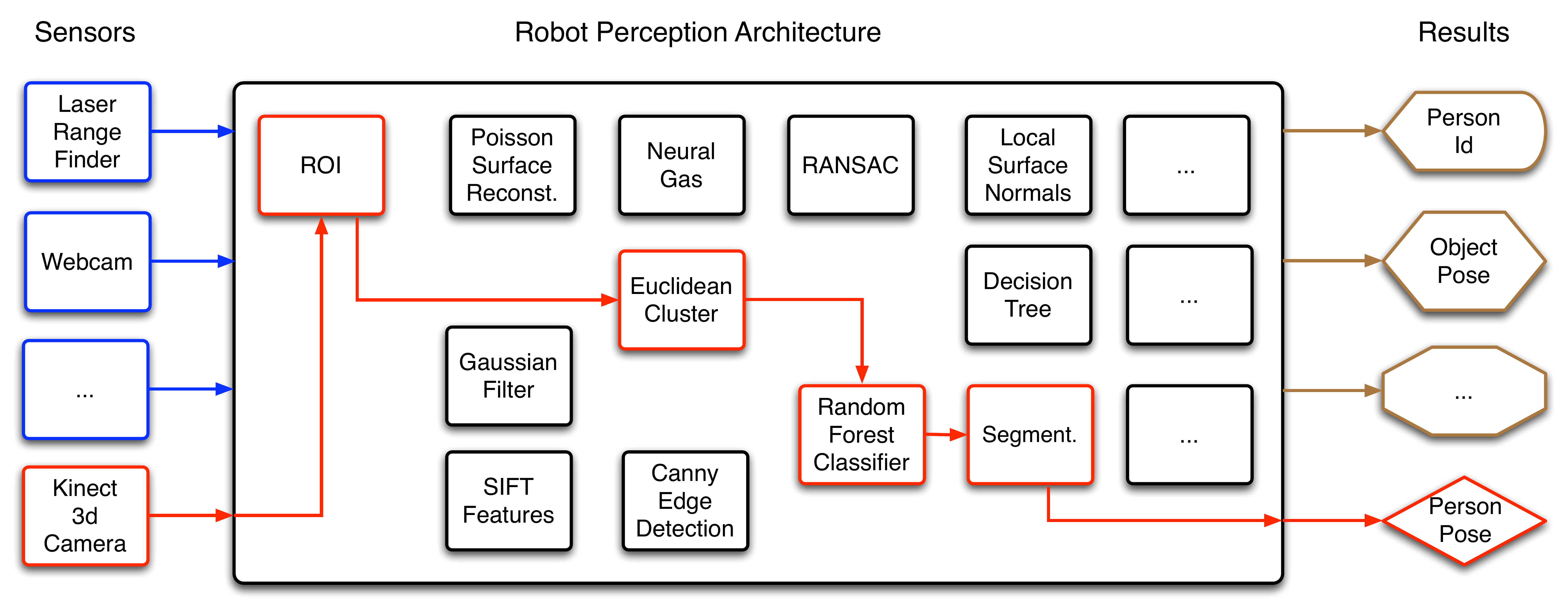}
  \caption{Elements making up the design space of a robot perception architecture: i) heterogenous sets of
sensors (blue boxes), ii) computational components (black boxes), and iii) task-relevant information
and knowledge (brown boxes). The path which is visualized in red shows an instance of an existing RPA described in Hegger \emph{et al.}~\cite{hegger2012}. The \emph{RPSL} is used
to specify the task knowledge visualized in brown boxes on the left hand side.}
  \label{fig:rpa}
\end{figure}

Domestic service robots such as PR2~\cite{PR2} and Care-O-bot 3\footnote{\url{http://www.care-o-bot-research.org/}} 
must be able to perform a wide range of different tasks ranging 
from opening doors~\cite{PR2} and making pancakes~\cite{pancakes11humanoids} to
serving drinks~\cite{Johnny}. A crucial precondition to achieve such complex tasks is the ability to
extract \emph{task knowledge} about the world from the data perceived through the sensors of the robot. 
Examples are the localization of humans~\cite{hegger2012} for navigation 
and interaction purposes, or the detection and recognition of objects in
images for the sake of manipulation by the robot.
To perceive all the knowledge needed to safely and robustly
perform a task, robots are equipped with a set of heterogenous sensors
such as laser range finders, Time-of-Flight (ToF) cameras, structured light
cameras and tactile sensors which provide different types of data
such as distance measurements, depth images, $3D$ point clouds, and
$2D$ grayscale or color images. To structure all the required processing steps
on this data so called \emph{Robot Perception Architectures} (RPAs) are 
required (see also Fig.~\ref{fig:rpa}). In general, RPAs are composed of functional 
components processing sensory input to output which is relevant for the task in hand. Thereby, heterogenous 
algorithms such as filters and feature detectors are integrated in components which 
are then assembled to make up an RPA~\cite{biggs2011}. However, despite recent algorithmic advancements in the field of vision and 
perception, the development of RPAs, designed to extract meaning out of 
the enormous amount of data, is still a complex and challenging exercise. 
There is little consensus on either how such an architecture is best designed for any 
particular task or on how to organize and structure robot perception architectures 
in general, so that they can accommodate the requirements for a \emph{wide range} of tasks. 

In this paper we present our work in progress towards a \emph{Robot Perception Specification 
Language (RPSL)}. \emph{RPSL} is a domain-specific language and its purpose is twofold. 
First, to provide means to specify the expected result (\emph{task knowledge}) of a 
RPA in an explicit manner\footnote{Referring to the right-hand side of Fig.~\ref{fig:rpa}}. Second, to initiate the (re)-configuration process of an RPA 
based on the provided specification. Here, we focus on the first objective and
discuss the core language concepts of \emph{RPSL}, namely the object, spatial, timing, dependency and composition domain.

\section{Problem Statement and Motivation}
\label{sec:problem}
Currently, robot perception architectures are developed by domain experts during design time. The design is significantly influenced by many decisions, which often remain
implicit. These design decision concern the robot \emph{platform}, the \emph{tasks} the robot should perform, and the \emph{environment} in which the robot operates.
Some exemplary design decisions include:
\begin{itemize}
	\item The sensor configuration (e.g., resolution or data frequency) of a particular sensor according to environment and task specifications.
	\item The general composition of an RPA, including the selection, configuration and organization of computational
components (implementing the core sensor processing functionalities such as filters, classifiers, etc.) such that
task and environment requirements are met.
	\item The configuration of a specific composition of an RPA for solving a particular task-relevant perception problem,
e.g. determining the pose of a human.
\end{itemize}
As long as task, environment, and platform specifications remain as assumed during design time, the RPA will operate
properly. However, if an event concerning robot capabilities, task requirements, or environment features occurs,
systematically ensuring an appropriate reaction by the RPA is challenging.
Generally speaking, the vast majority of RPAs is static and inflexible and it is not possible
\begin{itemize}
	\item to reconfigure parameters of computational components (e.g., the $\sigma$ value of a Gaussian filter) during run time,
	\item to execute complete processing chains in a demand-driven manner,
	\item and to modify and reconfigure robot perception processing chains during run time.
\end{itemize}
To provide RPAs with the ability to reconfigure their structure and behavior one needs to model the design decisions mentioned
above in an explicit and computable (during runtime available) manner. First of all the desired \emph{task knowledge} needs to 
be specified. Depending on the functional component (e.g., manipulation, grasping, or decision-making) which requires the knowledge
and the current task at hand this knowledge differs substantially. For instance, a decision-making component might be interested in
the existence of an object whereas a grasping and manipulation component demands more sophisticated information such
as spatial dimensions and shapes of an object. In both cases means to express the desired task knowledge are required. To the best of
our knowledge in robotics there is no language available which allows us to encode such specifications. We observe that often 
ad-hoc solutions e.g., in the form of message definitions (provided by the underlying robot software framework) are used which lack 
expressiveness and which are limited to certain frameworks.

\section{RPSL: Robot Perception Specification Language}
\label{sec:approach}
In the following we present the current status of the \emph{RPSL}. We identify first language requirements and then describe
the different domain concepts which are part of the language. Those concepts have been identified through
a domain analysis of existing RPAs and their application context in real-world scenarios.  

\subsection{Requirements and Assumptions}
The \emph{RPSL} is aimed to be a specification language. Therefore, the language is not executable. Interestingly, from a
planning point of view the specifications are comparable with goal specifications in the Planning Domain and Definition 
Language (PDDL)~\cite{pddl1998}. Similarly to PDDL a specification language for the perception domain should be 
independent of the underlying RPA just as PDDL is independent of concrete task planners. To be usable for a wide range 
of applications and systems, \emph{RPSL} should be independent of 
\begin{itemize}
	\item the type of sensor data processed by the RPA, and
	\item the type of functional components which are assembled to make up an RPA.
\end{itemize}
To enable reuse and exchangeability of the domain concepts realized in \emph{RPSL} (e.g., through concrete language primitives and abstraction) 
they should be orthogonal to each other as far as possible. Further, we assume that an environment is not actively observed (e.g., no 
active perception involving movements of the robot) due to the fact that required language concepts such as synchronization are not yet integrated in \emph{RPSL}. However, many so called 
table top situations in robotics are covered with the current status of the \emph{RPSL}.  

\subsection{General Approach}
Based on our domain analysis we derived several core domain concepts described in the following. To model the domains we apply a model-driven engineering
approach using the Eclipse Modeling Framework (EMF)~\cite{emf}. Here, each domain is specified in the form 
of an Ecore model. Based on the Ecore models we developed an external domain-specific language (DSL) with Xtext~\cite{xtext}. As the \emph{RPSL}
is work in progress we use the external DSL mainly to validate the domain concepts with experts. The next sections describe the domains and features
that need to be captured by the \emph{RPSL} in more detail.

\subsection{Object Domain}
\label{sec:objectdomain}
As exemplified in Fig.~\ref{fig:rpa} and discussed in Section~\ref{sec:problem} there is a huge variability
in the kind of \emph{task knowledge} potentially provided by RPAs. Ranging from diverse objects such 
as persons and objects of daily use such as cups, bottles, and door handles to the information about these
objects themself such as center of mass, poses, color and shapes. Here, the challenge is to use a representation 
which enables us to model the information about objects on various levels of abstraction. Ranging from raw sensor
data to feature descriptors and high-level object information such as size. In \emph{RPSL} the object 
domain is based on Conceptual Spaces (CS) which is a knowledge representation mechanism 
introduced by G\"{a}rdenfors~\cite{cs}. A conceptual space is composed of several (measurable) quality 
dimensions. A concept in a conceptual space is a convex region in that 
space. Points (also called knoxels) in a conceptual space represent concrete instances (objects) 
of a concept. To decide whether an instance belongs to one concept or to another we can apply 
similarity measures such as Euclidean distances. In Fig.~\ref{fig:exampleConceptRGB} an example
is shown. Here, a \texttt{Concept} called \emph{myBox} is specified. The concept belongs to the 
\texttt{Namespace} \emph{myConcepts} which is simply a mechanism to organize 
different concepts as known in general-purpose programming languages such as Java or C++.  
The concept \emph{myBox} uses the \texttt{Domain} \emph{Size} which is composed of three
quality dimensions, namely \emph{Height}, \emph{Width}, and \emph{Length}. In \emph{RPSL} quality dimensions 
with different scales such as continuous or ordinal scales are supported.  

\begin{figure}
\begin{center}
\tikzstyle{mybox} = [draw=gray, fill=white!20, very thick,
    rectangle, rounded corners, inner sep=2pt, inner ysep=2pt]
\begin{tikzpicture}
\node [mybox] (box){%
    \begin{minipage}{0.4\textwidth}
	  \begin{tabular}{l}
\begin{lstlisting}
myConcepts: Namespace {
	myBox: Concept {
		use_domain Size 
		p: Polytope {
			Point(Size.Height, 20mm)
			Point(Size.Height, 40mm)
			Point(Size.Width, 20mm)
			Point(Size.Width, 40mm)
			Point(Size.Length, 100mm)
		}	
	}
}
\end{lstlisting}
\end{tabular}
\end{minipage}
};
\end{tikzpicture}
\caption{Concept definition of a box.}
\label{fig:exampleConceptRGB}
\end{center}
\end{figure}

A \texttt{Polytope} is further used to model the ``borders'' of the concept \emph{myBox}. For instance, every box
belonging to the concept \emph{myBox} needs to have a height between $20mm$ and $40mm$. In contrast
to the Conceptual Space Markup Language (CSML) introduced by Adams and Raubal~\cite{csml} we use
polytopes instead of a set of inequalities to define the concept region as they are easier to 
model. To enrich the concept \emph{myBox} we simply refer to another domain. For instance, 
in Fig.~\ref{fig:exampleConcept} the concept \emph{myBox} is enriched with color information 
using the RGB color coding which includes three quality dimensions, namely, \emph{Red}, \emph{Green}, and \emph{Blue}. 
This approach allows us to model very expressive concepts as we can reuse existing domains and corresponding 
quality dimensions. Once concepts are defined we can model concrete instances or speaking
in the conceptual space terminology: ``prototypes''. In Fig.~\ref{fig:examplePrototype} a \texttt{Prototype} \emph{darkBlueBox} 
is modeled. Instead of defining ranges as in the concept definition, prototypes have single values per quality dimension. 

\begin{figure}
\begin{center}
\tikzstyle{mybox} = [draw=gray, fill=white!22, very thick,
    rectangle, rounded corners, inner sep=2pt, inner ysep=2pt]
\tikzstyle{fancytitle} =[fill= gray, text=white]
\begin{tikzpicture}
\node [mybox] (box){%
    \begin{minipage}{0.4\textwidth}
	  \begin{tabular}{l}
\begin{lstlisting}
myConcepts: Namespace {
	myBox: Concept {
		use_domain Size 
		use_domain RGB
		p: Polytope {
			// ...
			Point(RGB.Red, 0)
			Point(RGB.Green, 0)
			Point(RGB.Blue, 100)
			Point(RGB.Blue, 130)	
		}	
	}
}
\end{lstlisting}
\end{tabular}
\end{minipage}
};
\end{tikzpicture}
\caption{Concept definition of a box with color information.}
\label{fig:exampleConcept}
\end{center}
\end{figure}

\begin{figure}
\begin{center}
\tikzstyle{mybox} = [draw=gray, fill=white!22, very thick,
    rectangle, rounded corners, inner sep=2pt, inner ysep=2pt]
\tikzstyle{fancytitle} =[fill= gray, text=white]
\begin{tikzpicture}
\node [mybox] (box){%
    \begin{minipage}{0.4\textwidth}
	  \begin{tabular}{l}
\begin{lstlisting}
use Namespace myConcepts

darkBlueBox: Prototype {
	use_concept myConcepts.myBox
	v: Values {
		// ... 
		Point(myBox.RGB.Blue, 139)	
	}
}		
\end{lstlisting}
\end{tabular}
\end{minipage}
};
\end{tikzpicture}
\caption{Prototype definition of a dark blue box.}
\label{fig:examplePrototype}
\end{center}
\end{figure}

\begin{figure*}
\begin{center}
\tikzstyle{mybox} = [draw=gray, fill=white!22, very thick,
    rectangle, rounded corners, inner sep=2pt, inner ysep=2pt]
\tikzstyle{fancytitle} =[fill= gray, text=white]
\begin{tikzpicture}
\node [mybox] (box){%
    \begin{minipage}{0.8\textwidth}
	  \begin{tabular}{l}
\begin{lstlisting}
use Namespace myConcepts

boxSpec: Specification {
	d: Data {
		get Amount from myBox where myBox.Size.Width <= 20mm and myBox.Size.Length > 100mm
	}
}

darkBoxSpec: Specification {
	d: Data {
		get Amount from darkBlueBox where Similarity(EuclideanDistance) == 0
	}

darkBoxPoseSpec: Specification {
	d: Data {
		get Pose from darkBlueBox where Similarity(EuclideanDistance) == 0
	}
}

dependSpec: Specification {
	dg: DependencyGraph {
		darkBoxSpec before darkBoxPoseSpec 
	}
}

darkBoxDeadlineSpec: Specification {
	d: Data {
		get Amount from darkBlueBox where Similarity(EuclideanDistance) == 0 ensure Deadline(3s)
	}
}
\end{lstlisting}
\end{tabular}
\end{minipage}
};
\end{tikzpicture}
\caption{Some example specifications.}
\label{fig:examplespecs}
\end{center}
\end{figure*}

\subsection{Spatial Domain}
Very often it makes sense to specify the required object information with respect to the spatial surrounding. Assuming an egocentric view of the robot
one could model for instance objects through spatial operators such as ``behind'', ``next to'', and ``right of''. In particular, for manipulation 
tasks it is crucial to have information not only about the object to manipulate, but also about their spatial surrounding in order to plan motions
and to check for collisions. Currently, we investigate which spatial model we want to include in \emph{RPSL} such as the region connection calculus (RCC)~\cite{rcc}. 

\subsection{Timing Domain}
With the timing domain we intend to enrich specifications about the \emph{``when''}. More precisely, in many situations it is important to
retrieve information about objects within a certain time frame e.g., to avoid a stucking robot behavior. We use the notion of a deadline 
to encode a particular point in time by which the specified information should be available. For instance, specification \emph{darkBoxDeadlineSpec} shown 
in Fig.~\ref{fig:examplespecs} is enriched with a \texttt{Deadline} of $3s$. Here, \texttt{Deadline} can be parameterized with the value
and an time unit. From an implementation point of view once the specification
is received by the RPA it will obtain a time stamp which will be used to cope with the deadline. This imposes a certain protocol between
the component which emits the specification and the RPA which will not be discussed here. In future we intend to extend the timing domain
with policies allowing to model strategies with missed deadlines (e.g., ``when deadline X is missed try to retrieve information Y or 
repeat it once'').    

\subsection{Dependencies}
Another feature of the \emph{RPSL} is to model dependencies among specifications. That is some information
is required before some other information is available. In Fig.~\ref{fig:examplespecs}  specification
\emph{dependSpec} composed of two specifications which have a dependency. First, the amount of the \emph{darkBlueBox}
is retrieved and then the \emph{Pose} of the \emph{darkBlueBox} is retrieved. To model these situations the dependency 
meta-model is based on the concept of a  directed acyclic graph (DAG). Interestingly, in the past we used
the same dependency meta-model to model the sequence of component deployment~\cite{iros2013}. 

\subsection{Composition Domain}
The composition domain composes the previous domains in order to model a valid and complete specification. Some concepts such as timing and dependencies are optional whereas the object domain is mandatory. In Fig.~\ref{fig:examplespecs} some examples
are shown. First, the \texttt{Namespace} \emph{myConcepts} is used. Further, in the first specification one is interested in the amount of 
objects (visible in the current scene) belonging to the concept \emph{myBox} with certain properties concerning length and width. Here, \emph{Amount} 
itself is a concept with one quality dimension, namely an ordinal integer scale. As seen in the example
the syntax is inspired by SQL with the difference that the data model is based on Conceptual Spaces. Similarly to SQL we support logical operators
such as \texttt{AND} and \texttt{OR} as well as relational operators such as $==$, $>$ and $<=$ known from general-purpose programming languages. In the second
specification \emph{darkBoxSpec} the previously modeled prototype \emph{darkBlueBox} is used. After the \texttt{where} statement a condition is modeled. Here, the condition
is that only objects which look exactly like the \emph{darkBlueBox} (similarity measured with Euclidean distance) are counted. The idea is that with the \texttt{Similarity} operator
several similarity measures are supported and that we can balance the expected result according to the features provided by the measure. In doing so we will introduce also normalization methods for quality dimensions. In future we intend to support
also weighting factors which can be applied to increase or decrease the importance of quality dimensions for the similarity measure.

\section{Conclusion}
\label{sec:conclusions}
We presented the work in progress of using domain-specific
languages for specifying robot perception architectures. 
Assessing the DSLRob workshop series showed
that \emph{RPSL} is the first attempt to use DSLs 
in the sub-domain of robot perception. Even though, 
\emph{RPSL} is work in progress, it helped already to
identify and break down the crucial domains which 
are involved in specifying the result of RPAs. To 
achieve the second objective of our language, namely
the initialization of a (re)-configuration based on the specification
we are currently implementing a use case which is based
on simple table top scene.

\section*{Acknowledgement}
\small{Nico Hochgeschwender received a PhD scholarship from the Graduate Institute
of the Bonn-Rhein-Sieg University of Applied Sciences which is gratefully acknowledged. 
The authors also gratefully acknowledge the on-going support of the Bonn-Aachen International Center for 
Information Technology. Furthermore, the authors acknowledge the fruitful discussions at the 4th International 
Workshop on Domain-specific Languages and Models for ROBotics systems 
(DSLRob-13) co-located with IEEE/RSJ IROS 2013, Tokyo, Japan. Insights 
from the discussions have lead to an improved version of this paper.}




\bibliographystyle{IEEEtran}
\bibliography{proposal}

\end{document}